\documentclass{article}

\usepackage{tocloft}
\usepackage[preprint]{corl_2026}

\usepackage{amsmath,amssymb,amsfonts}
\usepackage{graphicx}
\usepackage{booktabs}
\usepackage{array}
\usepackage{multirow}
\usepackage{algorithm}
\usepackage{algorithmic}
\usepackage{subcaption}
\usepackage{hyperref}
\usepackage{cleveref}
\AtBeginDocument{%
  \setlength{\abovedisplayskip}{4pt plus 2pt minus 2pt}%
  \setlength{\belowdisplayskip}{4pt plus 2pt minus 2pt}%
  \setlength{\abovedisplayshortskip}{2pt plus 1pt}%
  \setlength{\belowdisplayshortskip}{2pt plus 1pt}%
}
\title{MemoryVAM: Integrating Memory into Video Action Model for Robot Manipulation}

\author{
  Yuxin Jiang$^{1*}$, Chang Yu$^{1*}$, Yunuo Chen$^{1}$, Xiang Feng$^{1,3}$, \\
  \textbf{Yin Yang$^{4}$, Nishank Gite$^{2}$, Chenfanfu Jiang$^{1}$}
  \\ $^{1}$University of California, Los Angeles, $^{2}$Nirvana Robotics, 
  \\ $^{3}$ University of California, San Diego, $^{4}$University of Utah
  \\ \footnotesize * Equal contributions.
}

\begin{document}
\maketitle

\vspace{-3.0em}
\begin{abstract}
Video-world-model policies learn action-relevant representations by predicting future observations. However, they condition on only a short observation window,
which renders long-horizon manipulation non-Markovian when the correct action depends on earlier events that are no longer visible.
We present \textbf{MemoryVAM}, an episodic memory mechanism for video-world-model policies.
We employ a Recap-Cue (RC) module, in which a Perceiver-based Recap Compressor maps per-frame CLIP embeddings into compact memory tokens, and a lightweight Cue Gate estimates task completion from memory and language.
These tokens are injected into both the video backbone and the action decoder, aligning policy imagination with episode progress and conditioning actions on history.
Our model trains the memory module with video prediction, a delta-reconstruction auxiliary loss, and episode-boundary supervision, requiring no per-frame progress labels.
The same mechanism applies to UNet and Diffusion Transformer (DiT) backbones by changing only the cross-attention injection interface.
On LIBERO-Mem, our model improves average success from 5\% to 42.5\%.
On real robots, it achieves 78.3\% success on counting tasks, 80.0\% on spatial recall, and 75.0\% on sequential tracking. Project page: \url{https://MemoryVAM.github.io/}.
\end{abstract}


\addtocontents{toc}{\protect\setcounter{tocdepth}{-1}}

\begin{figure}[t]
    \centering
    \includegraphics[width=\linewidth]{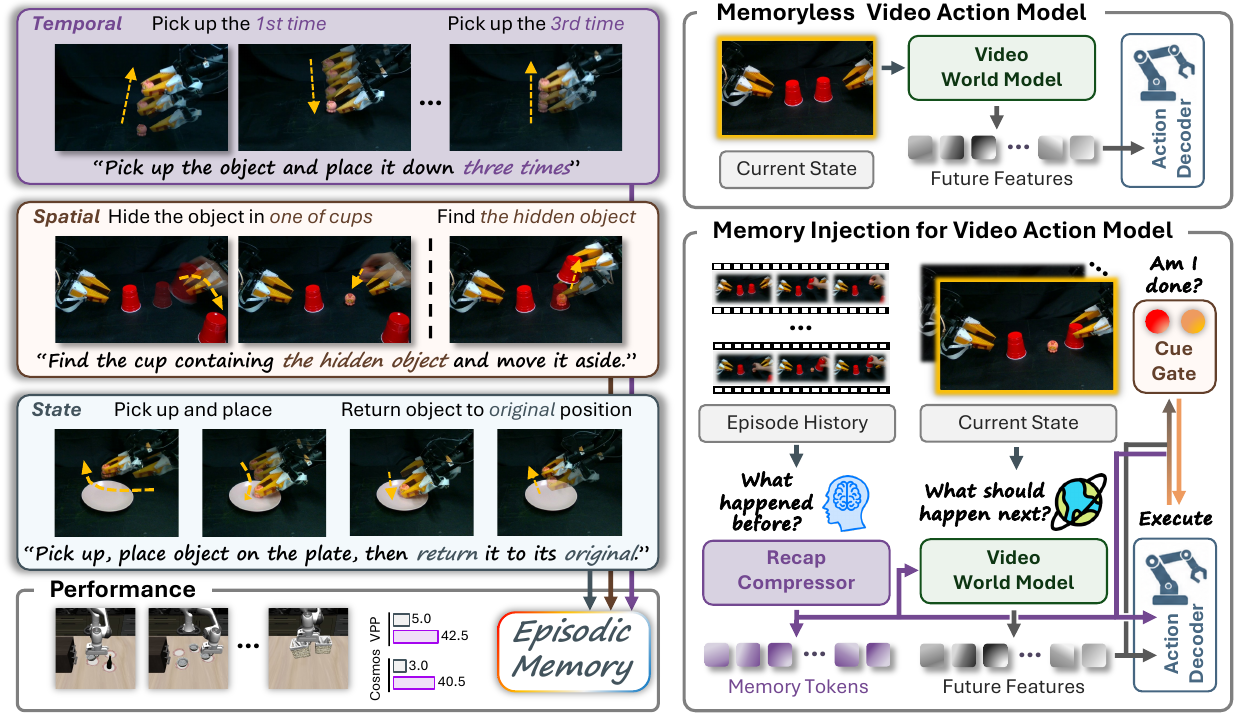}
    \caption{\textbf{Method Overview.}
    Memory-dependent manipulation requires recalling temporal progression, latent state, and spatial history. In contrast to memoryless world-action models, which condition only on the current state, we introduce episode-level memory to enable history-aware prediction, task completion, and action decoding. This mechanism improves both VPP~\citep{du2024vpp} and Cosmos-RF~\citep{agarwal2025cosmos,kim2026cosmos} backbones on LIBERO-Mem~\citep{slotssmv2} and physical robot manipulation tasks.
    }
    \label{fig:concept}
    \vspace{-8pt}
\end{figure}

\section{Introduction}
\label{sec:intro}

Video world models provide a useful representation source for robot policies by learning how visual observations evolve over time~\citep{hou2026world}. For a robotic action policy, such features allow the decoder to condition on future changes in addition to the current observation. Recent video-based robot policies instantiate this idea with different backbone architectures. VPP~\citep{du2024vpp} uses a fine-tuned Stable Video Diffusion (SVD) UNet model as the video predictor, while Diffusion Transformer (DiT)-based Cosmos video generators~\citep{agarwal2025cosmos} have shown success for action prediction in MimicVideo~\citep{pai2025mimicvideo}, Genie Envisioner~\citep{liao2025genieenvisioner}, and Cosmos Policy~\citep{kim2026cosmos}. Across these systems, the shared design choice is to use a predictive video backbone as the visual representation source for downstream action prediction, which is effective when tasks are approximately Markovian: the current observation is sufficient to infer the relevant future dynamics.

While video models provide strong predictive capabilities for policy learning, they typically rely on a fixed-length observation window for both future prediction and action decoding. The video backbone predicts future frames from recent visual context and language, while the action decoder uses features derived from the same limited window. This design is particularly limiting in long-horizon, multi-stage tasks where visually similar states can require different actions. For example, in the task \emph{``pick up the red block three times,''} frames corresponding to different repetitions may appear nearly identical, while the correct next action depends on hidden task progress. Under a fixed observation window, such tasks become non-Markovian without episodic memory.

This memory issue is not unique to video-world-model policies. Recent work studies it in diffusion-policy settings with Gated Memory Policy~\citep{gao2026gatedmemorypolicy} and in Vision-Language-Action (VLA) models, including MemoryVLA~\citep{memoryvla2025}, MemER~\citep{memer2026}, and $\pi_{\text{mem}}$~\citep{torne2026mem}. For video-world-model policies, the distinctive aspect is that memory can be supervised by the generative backbone. Dense future-frame prediction allows memory to serve as a conditioning signal for what the model should imagine next, rather than solely as an input to the action head. The action decoder then consumes backbone features that already encode episode progress, thereby injecting history awareness to subsequent action prediction.
A video-world-model policy therefore requires memory awareness in both components: the backbone should generate futures consistent with episode history, and the action decoder should select controls conditioned on the same memory state.

We present \textbf{MemoryVAM}, which adds episodic memory to video-world-model policies through dual memory conditioning (\Cref{fig:concept}). Our method enables episode history to shape both the video backbone’s future predictions and the action decoder’s control decisions, so that memory both supervises future-state imagination and directly constrains policy execution.
Our contributions are as follows: (1) We introduce \emph{dual injection}, a two-pathway memory interface that integrates compact Recap tokens into both the video backbone and the action decoder, allowing history to jointly influence imagined futures and control features. (2) We design the \emph{Recap-Cue (RC)} module, which combines a \emph{Recap Compressor} trained with video and reconstruction objectives to produce compact yet informative memory representations, and a \emph{Cue Gate} trained from episode boundaries to enable autonomous task completion. (3) We instantiate this mechanism across both UNet and DiT video backbones, demonstrating that our method is not tied to a specific backbone paradigm.
On LIBERO-Mem~\citep{slotssmv2}, our model improves average success from 5\% to 42.5\% on tasks requiring episodic recall. On real robots, it achieves 78.3\% success on counting tasks, 80.0\% on spatial recall, and 75.0\% on sequential tracking.

\section{Related Work}
\label{sec:related}

\paragraph{Video World Models for Robot Learning.}
Video world and action models benefit robot policies by learning how current observations connect to likely future scene changes and feasible actions. We refer readers to \citep{hou2026world} for a comprehensive survey. One branch uses generated video as a policy interface or subgoal signal, including UniPi~\citep{du2023unipi}, SuSIE~\citep{black2024susie}, and GR-1~\citep{wu2024gr1}. Another branch uses video backbones as action-relevant representations. Video Prediction Policy (VPP) extracts intermediate Stable Video Diffusion (SVD) UNet features~\citep{du2024vpp}, while Diffusion Transformer (DiT) systems include Genie~\citep{bruce2024genie}, Genie Envisioner~\citep{liao2025genieenvisioner}, Cosmos~\citep{agarwal2025cosmos}, Cosmos Policy~\citep{kim2026cosmos}, and MimicVideo~\citep{pai2025mimicvideo}. Other video-generation formulations, including Diffusion Forcing~\citep{chen2024diffusionforcing} and Self-Forcing~\citep{huang2025selfforcing}, improve sequence prediction and planning objectives. However, these methods typically condition only on recent frames rather than episode-level memory, whereas our approach demonstrates that explicit episodic memory is crucial for long-horizon tasks across diverse video backbone architectures.

\paragraph{Memory Mechanisms for Robot Policies.}
Prior work studies memory through benchmarks, history compression, and policy-level memory. LIBERO~\citep{liu2023libero} introduced memory-dependent benchmarks, and MA-LMM~\citep{he2024malmm} compresses long video histories for vision-language models. DeltaTok~\citep{kerssies2026deltatok} shows that self-supervised temporal compression can preserve video changes, while Perceiver-style resamplers~\citep{jaegle2021perceiver,jaegle2022perceiverio}, Flamingo~\citep{alayrac2022flamingo}, and Q-Former~\citep{li2023blip2} map context into fixed-size latent tokens. In robot policy learning, Gated Memory Policy~\citep{gao2026gatedmemorypolicy} adds memory to diffusion policies, and VLA-oriented methods such as MemER~\citep{memer2026}, MemoryVLA~\citep{memoryvla2025}, $\pi_{\text{mem}}$~\citep{torne2026mem}, and RoboMME~\citep{robomme2026} study memory for action. In this paper, we focus on video world model policies and investigate whether episode-level memory should be incorporated into the predictive representation used for both future prediction and action selection. This perspective treats memory as an integral component of the world-model state, rather than merely an auxiliary input to the action policy.

\section{Method}
\label{sec:method}

\begin{figure}[t]
    \centering
    \includegraphics[width=\linewidth]{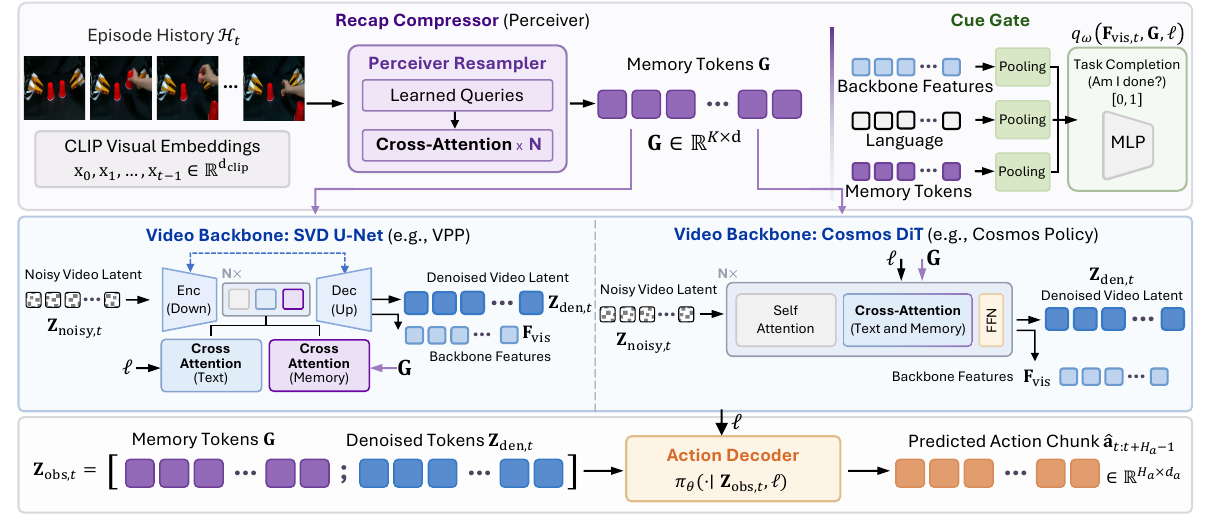}
    \caption{\textbf{Architecture Overview.} Episode history (Sec.~\ref{sec:method:history}) is first compressed by the Recap Compressor (Sec.~\ref{sec:method:resampler}) into compact memory tokens, which are injected through two complementary pathways (Sec.~\ref{sec:method:dual_injection}): into the Video Backbone to make future prediction history-aware, and into the Action Decoder to condition control on the same episodic memory. The Cue Gate further uses backbone features, memory tokens, and language to estimate task completion (Sec.~\ref{sec:method:eos}).}
    \label{fig:architecture}
    \vspace{-1.0em}
\end{figure}

At timestep $t$, let $\ell$ denote the language instruction, $\mathbf{I}_{t-r+1:t}$ the local history of length $r$, and $\mathbf{G}$ the memory tokens produced from the episode history. The video world model $p_\phi$ predicts an $H_v$-step future visual sequence, and its feature extractor $f_\phi$ provides intermediate features for control,
\[
\hat{\mathbf{I}}_{t+1:t+H_v} \sim p_\phi(\cdot \mid \mathbf{I}_{t-r+1:t}, \ell, \mathbf{G}), \qquad
\mathbf{Z}_{\text{den},t} = f_\phi(\mathbf{I}_{t-r+1:t}, \ell, \mathbf{G}).
\]
The action policy $\pi_\theta$ takes observation tokens $\mathbf{Z}_{\text{obs},t}$ and predicts an $H_a$-step action chunk $\hat{\mathbf{A}}_t$, where $d_a$ is the action dimension,
\[
\mathbf{Z}_{\text{obs},t} = [\mathbf{G};\mathbf{Z}_{\text{den},t}], \qquad
\hat{\mathbf{A}}_t = \hat{\mathbf{a}}_{t:t+H_a-1} \sim \pi_\theta(\cdot \mid \mathbf{Z}_{\text{obs},t}, \ell), \qquad
\hat{\mathbf{A}}_t \in \mathbb{R}^{H_a \times d_a}.
\]
We realize these two uses of memory through complementary pathways (\Cref{fig:architecture}). \emph{Pathway~1} injects $\mathbf{G}$ into the video model so that predicted futures and backbone features $\mathbf{F}_{\text{vis}}$ are conditioned on episode history. \emph{Pathway~2} exposes the same $\mathbf{G}$ to the action decoder through $\mathbf{Z}_{\text{obs},t}$. A lightweight Cue Gate estimates $q_\omega(\mathbf{F}_{\text{vis}}, \mathbf{G}, \ell)$ for task completion and autonomous termination.

\subsection{Episode History Encoding}
\label{sec:method:history}

At each timestep $t$, we maintain an episode history $\mathcal{H}_t = \{\mathbf{x}_0, \ldots, \mathbf{x}_{t-1}\}$, where $\mathbf{x}_i = \mathrm{CLIP}_{\mathrm{vis}}(\mathbf{I}_i) \in \mathbb{R}^{d_{\text{clip}}}$ is the CLIP visual embedding of frame $\mathbf{I}_i$~\citep{radford2021learning}. CLIP's language-aligned visual features let the memory track task-relevant semantic changes across the episode rather than low-level pixel differences alone.

\subsection{Recap Compressor}
\label{sec:method:resampler}

The Recap Compressor transforms the variable-length episode history $\mathcal{H}_t$ into a fixed-size memory matrix $\mathbf{G} \in \mathbb{R}^{K \times d}$ via \emph{Perceiver} cross-attention~\citep{jaegle2021perceiver,jaegle2022perceiverio}, following the Flamingo resampler architecture~\citep{alayrac2022flamingo}. 
Rather than uniformly pooling the full history, learned queries selectively attend to events that explain task progress, providing a stable memory interface to both the video backbone and the action decoder. Order-aware positional encoding enables $\mathbf{G}$ to capture not only which salient events occurred, but also when they occurred within histories of up to $T_{\text{hist}}$ frames.

\subsection{Dual Memory Conditioning}
\label{sec:method:dual_injection}

The central mechanism of our method is dual injection: the same memory tokens $\mathbf{G}$ make both future prediction and action decoding history-aware. Pathway~1 aligns imagined futures and their corresponding backbone features with episode progress, while Pathway~2 constrains the action distribution using the same memory.

\paragraph{Pathway 1: Cross-attention injection into the video backbone.}
Pathway~1 treats $\mathbf{G}$ as an additional conditioning matrix for the video backbone's cross-attention layers. For a cross-attention layer $m$ with hidden states $\mathbf{H}^{(m)}$ and text-conditioning matrix $\mathbf{C}_{\text{text}}^{(m)}$, we adopt a backbone-specific interface while keeping the Recap output fixed. For the Cosmos DiT backbone~\citep{agarwal2025cosmos}, memory tokens are appended to the conditioning matrix:
\[
\tilde{\mathbf{C}}^{(m)} = [\mathbf{C}_{\text{text}}^{(m)};\ \mathbf{G}], \qquad
\mathbf{O}^{(m)} = \operatorname{CrossAttn}^{(m)}(\mathbf{H}^{(m)}, \tilde{\mathbf{C}}^{(m)}).
\]
For the SVD UNet backbone used by VPP~\citep{du2024vpp}, we retain the text cross-attention branch and introduce a separate memory branch with layer-specific projections, following decoupled cross-attention~\citep{ye2023ipadapter}:
\[
\mathbf{O}^{(m)} =
\operatorname{CrossAttn}_{\text{text}}^{(m)}(\mathbf{H}^{(m)}, \mathbf{C}_{\text{text}}^{(m)})
+
\operatorname{CrossAttn}_{\text{mem}}^{(m)}(\mathbf{H}^{(m)}, \mathbf{G}).
\]
Thus, the same memory tokens $\mathbf{G}$ condition both backbone families through their cross-attention, with backbone-specific differences confined to how $\mathbf{G}$ is mapped into each conditioning pathway.

\paragraph{Pathway 2: Direct injection into action policy.}
Pathway~2 conditions the action policy directly on the same memory matrix $\mathbf{G}$ used by the video backbone. For UNet-backbone policies with token-input action decoders, let $\mathbf{Z}_{\text{den},t} \in \mathbb{R}^{N_{\text{den}} \times d}$ denote denoised video tokens extracted from backbone features. The action decoder receives the observation-token matrix
\[
\mathbf{Z}_{\text{obs},t} = [\mathbf{G};\ \mathbf{Z}_{\text{den},t}] \in \mathbb{R}^{N_{\text{obs}} \times d}, \qquad
N_{\text{obs}} = K + N_{\text{den}}.
\]
The resulting action distribution $\pi_\theta(\cdot \mid \mathbf{Z}_{\text{obs},t}, \ell)$ depends on episode history through the explicit memory tokens and through the Pathway~1 features contained in $\mathbf{Z}_{\text{den},t}$. For DiT-backbone policies with cross-attention action transformers, let $\mathbf{S}_t^{(n)}$ be the action hidden states at layer $n$ and $\mathbf{C}_{\text{vis},t}^{(n)}$ be the visual conditioning matrix derived from DiT features. With a layer-specific memory projection $\rho_n$, Pathway~2 forms
\[
\tilde{\mathbf{C}}_{\text{act},t}^{(n)} = [\mathbf{C}_{\text{vis},t}^{(n)};\ \rho_n(\mathbf{G})], \qquad
\mathbf{S}_t^{(n+1)} = \operatorname{CrossAttn}_{\text{act}}^{(n)}(\mathbf{S}_t^{(n)}, \tilde{\mathbf{C}}_{\text{act},t}^{(n)}).
\]
Both interfaces make the action distribution history-conditioned while keeping the same Recap memory tokens across UNet and DiT backbone families.

\subsection{Cue Gate}
\label{sec:method:eos}

The Cue Gate converts episodic memory into an autonomous stopping signal. It estimates the completion score $q_\omega(\mathbf{F}_{\text{vis},t}, \mathbf{G}, \ell) \in [0,1]$ from pooled backbone features, pooled memory tokens, and the language instruction. This matters for memory-dependent tasks because completion may depend on hidden episode progress rather than the current frame alone. Supervision comes directly from demonstrations, where final frames are positive and earlier frames are negative, so no per-step progress labels are needed. Beyond test-time termination, this objective provides a self-supervised task-level progress signal for the memory module, encouraging the Recap Compressor to retain information that is useful for both deciding when the goal is complete and selecting the next action. Appendix~\ref{app:progress_gate} provides the full architecture.

\subsection{Backbone Instantiations}
\label{sec:method:backbone_injection}

To effectively inject history-awareness, we design a unified memory interface for two-stage video action models. In this paradigm, a video backbone predicts future observations and exposes visual features, and a downstream action module maps those features to controls. We validate the same interface on two representative instantiations: VPP~\citep{du2024vpp}, which uses a UNet video backbone, and Cosmos-RF, which uses a Diffusion Transformer video backbone based on Cosmos~\citep{agarwal2025cosmos,kim2026cosmos} and follows the action-decoding setup of MimicVideo~\citep{pai2025mimicvideo}. Across both cases, we change the information available to the two-stage policy rather than introducing backbone-specific memory modules, so the same Recap Compressor supports history-aware future prediction and action decoding.

\subsection{Training}
\label{sec:method:training}

We use a two-stage training procedure to separate history-aware representation learning from control learning. Stage~1 teaches the video backbone and memory module to predict futures that respect episode history, using the video prediction objective together with the reconstruction auxiliary loss and the Cue Gate loss. Stage~2 trains the action policy on the resulting memory-conditioned features while keeping the backbone stable, so the controller learns from a consistent predictive representation. Optimizer settings, training schedules, reconstruction details, and inference mechanics are deferred to Appendix~\ref{app:training_details}.

\begin{figure}[t]
    \centering
    \includegraphics[width=\linewidth]{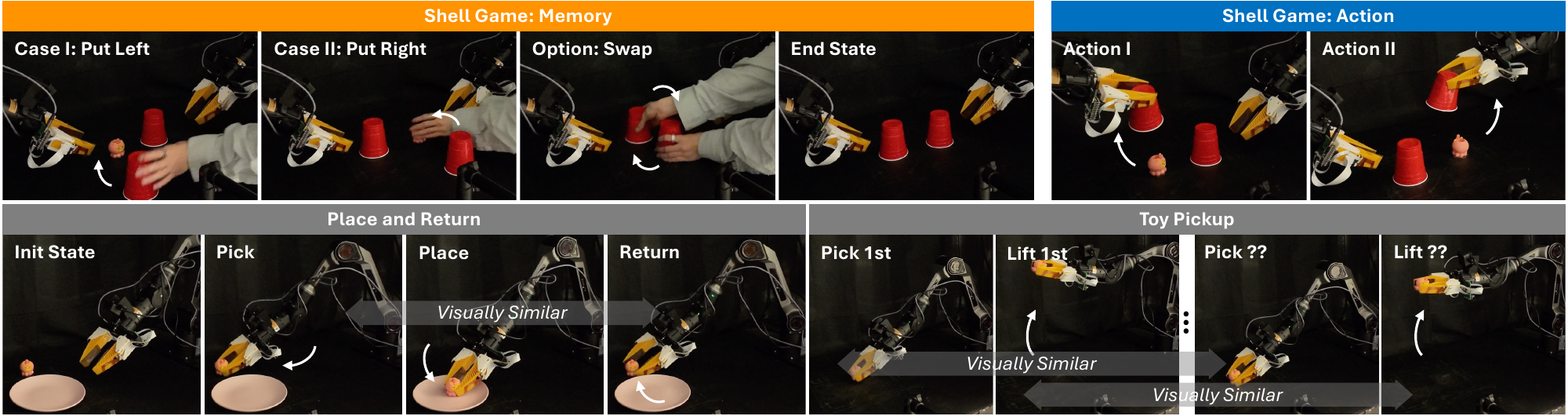}
    \caption{\textbf{Real-Robot Task Settings.} In \emph{Shell Game}, the toy is placed under a cup, the cups may be swapped, and the robot must later lift the cup containing the hidden toy. \emph{Place and Return} requires the robot to remember the initial object position after moving the object to the plate. \emph{Toy Pickup $k\times$} tests counting, since later pick-and-place repetitions can begin from visually similar states.}
    \vspace{-1.0em}\label{fig:real_robot}
\end{figure}

\section{Experiments}
\label{sec:exp}

We evaluate our method with the objective of determining whether episodic memory improves both the predictive backbone and the downstream robot policy. We first assess video prediction quality in memory-dependent scenarios (Sec.~\ref{sec:exp:vidpred}). Next, we evaluate policy performance on the LIBERO-Mem~\citep{slotssmv2} benchmark, comparing against memoryless baselines and memory-augmented VLA and diffusion-policy methods (Sec.~\ref{sec:exp:libero}), as well as on three real-robot tasks that require recalling earlier episode events (Sec.~\ref{sec:exp:real}). Finally, we ablate the reconstruction loss and termination behavior to identify the components driving the observed gains (Sec.~\ref{sec:exp:ablation}).

\subsection{Experimental Setup}
\label{sec:exp:setup}

\paragraph{Backbones.}
We evaluate our method on two representative video backbone families: the UNet-based VPP~\citep{du2024vpp} and the DiT-based Cosmos-RF built upon Cosmos~\citep{agarwal2025cosmos,kim2026cosmos}. Each no-memory baseline retains the same backbone and action module while removing episodic memory, thereby isolating the contribution of memory injection.

\paragraph{Baselines and variants.}

For each video backbone, we compare the memoryless baseline with the full memory-injected model. On VPP, we further report single-path ablations that retain only Pathway~1 or Pathway~2, isolating the roles of memory in the predictive backbone and the action decoder. External baselines include memoryless VLA references $\pi_0$~\citep{black2024pi0} and SlotVLA~\citep{slotssmv2}; memory-augmented VLA methods SlotVLA-with-history~\citep{slotssmv2}, Embodied-SlotSSM~\citep{slotssmv2}, and MemoryVLA~\citep{memoryvla2025}; and the memory-augmented diffusion-policy method Gated Memory Policy~\citep{gao2026gatedmemorypolicy}. Embodied-SlotSSM uses oracle subgoal information, and all trainable baselines are fine-tuned on the same set of demonstrations.
More details on architecture, training settings, and parameter values are provided in Appendices~\ref{app:implementation}, \ref{app:training_details}, and~\ref{app:architecture_parameters}.

\paragraph{Tasks.}
We evaluate on four task domains: (1) LIBERO-Mem~\citep{slotssmv2}, a simulation benchmark comprising 10 memory-dependent manipulation tasks spanning counting, sequencing, and spatial reasoning (\Cref{tab:libero}); (2) \emph{Toy Pickup $k\times$}, where the robot repeatedly performs pick-and-place for $k \in \{1, 3, 5\}$ trials to test counting ability; (3) \emph{Place and Return}, where the robot must restore an object to its initial position after an intermediate placement; and (4) \emph{Shell Game}, where the robot tracks an occluded object after cup swapping. The real-robot setup is described in Appendix~\ref{app:real_robot_setup}.

\subsection{Video Prediction Quality}
\label{sec:exp:vidpred}

We first evaluate whether memory injection improves the video model's prediction quality. Following the evaluation protocol of Stable Video Diffusion~\citep{blattmann2023svd}, we measure the final frame of an autoregressive rollout against the ground-truth frame, since each predicted chunk is fed back as context and later frames accumulate rollout errors. We also evaluate iterative prediction, where the model repeatedly predicts an $H_v$-frame chunk until reaching horizons of $2H_v$, $4H_v$, and $8H_v$.

\paragraph{Metrics.}
We report Last-Frame SSIM $\uparrow$ and Last-Frame LPIPS $\downarrow$ on the final rollout frame, together with FVD $\downarrow$ at autoregressive horizons $H_v$, $2H_v$, $4H_v$, and $8H_v$.
On memory-dependent tasks, pixel realism alone is insufficient because a video can look plausible while representing the wrong task progress, such as repeating a completed sub-task. These metrics therefore test whether the rollout remains visually faithful and semantically aligned with episode history.

\begin{figure}[t]
    \centering
    \includegraphics[width=.9\linewidth]{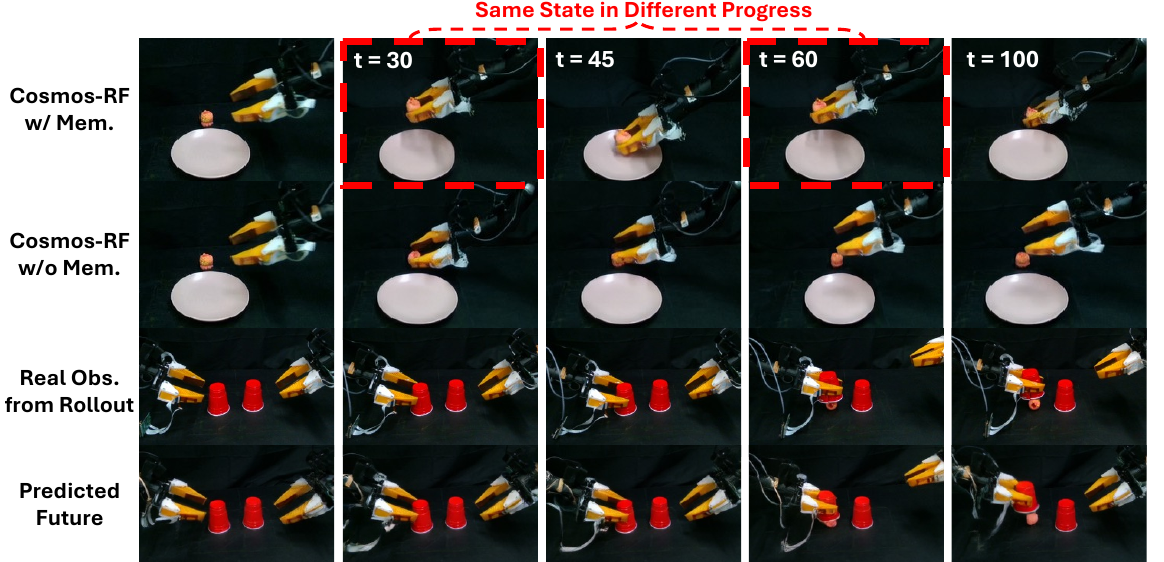}
    \caption{\textbf{Qualitative Results of Memory Injection}. \emph{Top}: Cosmos-RF w/ vs. w/o memory, showing that memory enables the policy to recall completed states. \emph{Bottom}: predicted future frames vs. real observations during real-robot deployment, demonstrating faithful visual prediction.}
    \vspace{-1.0em}\label{fig:video_gen_quality}
\end{figure}

\Cref{fig:video_gen_quality} shows that episodic memory primarily improves history alignment during rollout. Compared to VPP, our UNet-based model improves both single-chunk fidelity and iterative prediction, reducing FVD from 43.36 to 18.32 at $H_v$ and from 37.81 to 19.84 at $8H_v$. Compared to Cosmos-RF, our DiT-based model improves single-chunk FVD from 22.54 to 18.65 and long-horizon rollout FVD from 26.11 to 16.24 at $8H_v$. This pattern suggests that memory is most beneficial when predictions must remain aligned with hidden task progress over long autoregressive horizons.

\subsection{Main Benchmark Results}
\label{sec:exp:libero}

\begin{table}[t]
    \centering
    \caption{\textbf{LIBERO-Mem Results.} Success rate (\%) over 20 rollouts per task. Columns cover single placement, counting, and sequencing/spatial tasks. Abbreviations: B=Bowl, Bt=Bottle, P=plate, Cs=Cheese, Bskt=basket.}
    \label{tab:libero}
    \small
    \setlength{\tabcolsep}{2.8pt}
    \begin{tabular}{@{}l cc cccc cccc c@{}}
        \toprule
        & \multicolumn{2}{c}{Single} & \multicolumn{4}{c}{Counting} & \multicolumn{4}{c}{Sequencing / Spatial} & \\
        \cmidrule(lr){2-3} \cmidrule(lr){4-7} \cmidrule(lr){8-11}
        Method & B$\to$P & Bt$\to$P & B$\times$3 & Bt$\times$3 & B$\times$5 & B$\times$7 & Swap2 & Rot3 & Cs$\to$Bskt & Cs+E & Avg \\
        \midrule
        \multicolumn{12}{l}{\textit{VPP backbone (SVD UNet, 1.5B)}} \\
        No memory          & 30.0          & 10.0          & 10.0          & 0.0           & 0.0           & 0.0  & 0.0           & 0.0          & 0.0           & 0.0           & 5.0  \\
        Pathway 1 only            & \textbf{85.0}      & 40.0      & 35.0      & 10.0      & 15.0      & 0.0 & 5.0   & 0.0     & 30.0      & 65.0      & 28.5 \\
        Pathway 2 only            & 45.0      & 10.0      & 5.0      & 0.0      & 5.0      & 5.0 & 0.0   & 0.0     & 5.0      & 0.0      & 7.5 \\
        \textbf{Ours (UNet)}    & \textbf{85.0} & \textbf{65.0} & \textbf{90.0} & 10.0 & 35.0 & 0.0  & \textbf{25.0} & 5.0 & \textbf{40.0} & \textbf{70.0} & \textbf{42.5} \\
        \midrule
        \multicolumn{12}{l}{\textit{Cosmos-RF backbone (rectified-flow, 2B)}} \\
        No memory          & 0.0           & 5.0           & 0.0           & 0.0           & 0.0           & 0.0  & 0.0           & 0.0          & 15.0          & 10.0          & 3.0  \\
        \textbf{Ours (DiT)}    & 75.0 & \textbf{65.0} & 65.0 & \textbf{15.0} & \textbf{55.0} & \textbf{35.0} & 10.0 & \textbf{10.0} & 25.0 & 50.0 & 40.5 \\
        \midrule
        \multicolumn{12}{l}{\textit{VLA baselines, no memory}} \\
        $\pi_0$~\citep{black2024pi0}              & 50.0 & 0.0  & 0.0  & 0.0  & 0.0  & 0.0  & 0.0  & 0.0  & 0.0  & 0.0  & 5.0  \\
        SlotVLA~\citep{slotssmv2}              & 0.0  & 0.0  & 0.0  & 0.0  & 0.0  & 0.0  & 0.0  & 0.0  & 0.0  & 0.0  & 0.0  \\
        \midrule
        \multicolumn{12}{l}{\textit{VLA baselines, memory-augmented}} \\
        SlotVLA$_{h=8}$~\citep{slotssmv2}      & 50.0 & 0.0  & 0.0  & 0.0  & 0.0  & 0.0  & 0.0  & 0.0  & 0.0  & 0.0  & 5.0  \\
        E-SlotSSM~\citep{slotssmv2}            & 50.0 & 0.0  & 35.0 & 0.0  & 15.0 & 0.0  & 0.0  & 0.0  & 30.0 & 20.0 & 15.0 \\
        MemoryVLA~\citep{memoryvla2025}        & 55.0 & 0.0 & 35.0 & 0.0 & 5.0 & 0.0 & 0.0 & 0.0 & 0.0 & 0.0 & 9.5 \\
        \midrule
        \multicolumn{12}{l}{\textit{Diffusion-policy baseline, memory-augmented}} \\
        Gated Memory Policy~\citep{gao2026gatedmemorypolicy} & \textbf{85.0} & 45.0 & 20.0 & 5.0  & 40.0 & 0.0  & 0.0  & 0.0  & 0.0  & 0.0  & 19.5 \\
        \bottomrule
    \end{tabular}
\end{table}

\Cref{tab:libero} illustrates that the main challenge lies in recovering hidden task progress rather than executing isolated pick-and-place motions. Memoryless policies can occasionally solve single-step variants, but performance drops sharply when visually similar scenes correspond to different counts, object histories, or sequence states. Memory injection improves performance on both VPP and Cosmos-RF, indicating that the mechanism is not tied to a specific video backbone family. Compared with memory-augmented VLA and diffusion-policy baselines, the gains are most pronounced on tasks where memory must be grounded in the future scene state that the robot is expected to realize. This suggests that episodic memory should shape not only the policy state but also the predictive representation from which actions are decoded.

\begin{table}[b]
    \vspace{-2em}
    \centering
    \caption{\textbf{Real-Robot Results.} Success rate (\%) across real-robot memory tasks.}
    \label{tab:real_robot}
    \small
    \begin{tabular*}{\linewidth}{@{\extracolsep{\fill}}l cccc c c@{}}
        \toprule
        & \multicolumn{4}{c}{\emph{Toy Pickup $k\times$}} & \emph{Place and Return} & \emph{Shell Game} \\
        \cmidrule(lr){2-5} \cmidrule(lr){6-6} \cmidrule(lr){7-7}
        Method & $1\times$ & $3\times$ & $5\times$ & Avg & & \\
        \midrule
        Cosmos-RF (no mem) & 0.0 & 0.0 & 0.0 & 0.0 & 0.0 & 5.0 \\
            \textbf{Ours (DiT)} & \textbf{95.0} & \textbf{100.0} & \textbf{40.0} & \textbf{78.3} & \textbf{80.0} & \textbf{75.0} \\
        \bottomrule
    \end{tabular*}%
    \vspace{-1.0em}
\vspace{-1.0em}
\end{table}

\subsection{Real Robot Experiments}
\label{sec:exp:real}

We evaluate our model on three real-robot tasks whose protocols are shown in \Cref{fig:real_robot}. The physical setup is shown in \Cref{fig:robot_setup}, with hardware details in Appendix~\ref{app:real_robot_setup}. Our history-aware policy achieves 78.3\% average success on \emph{Toy Pickup $k\times$}, 80.0\% on \emph{Place and Return}, and 75.0\% on \emph{Shell Game}.
The common stressor is observation aliasing in the physical world. At key decision points, the current image is ambiguous because multiple episode states can produce similar observations. The correct action may depend on a repetition count, a previous object pose, or occluded object information. This setup tests whether episodic memory remains useful outside simulation, where visual variation and execution noise make progress inference harder.
\Cref{tab:real_robot} follows the same trend as LIBERO-Mem. The memoryless Cosmos-RF policy collapses when task progress is not observable, while our memory-injected policy achieves 78.0\% average success across five real-robot evaluation settings. These results indicate that the performance gains transfer to physical rollouts, where history is essential for disambiguating the next action.

\subsection{Ablation Studies}
\label{sec:exp:ablation}

We conduct ablations on the VPP backbone using LIBERO-Mem~\citep{slotssmv2}.

\paragraph{Reconstruction loss and completion timing.}
The completion-timing results in \Cref{tab:eos_ablation} measure whether the Cue Gate predicts task completion at the right time.
The reconstruction auxiliary loss reduces mean timing error from 44.3 to 35.5 frames at $p \geq 0.5$, with the largest gains on repetitive tasks where counting matters.
At the stricter threshold $p \geq 0.9$, the gain is larger, improving from 62.0 to 28.8 frames. This suggests that reconstruction does more than shift the average trigger time: it produces a sharper and better-calibrated completion signal, which is important when termination depends on remembered task progress rather than the current frame alone.

\paragraph{Video generation termination quality.}
We evaluate termination in the generated video because a world model that understands task progress should not keep imagining unnecessary motion after the goal is complete. We report stop ratio, defined as post-completion motion divided by pre-completion motion in \Cref{app:stop}, with results shown in \Cref{tab:video_stop}.
Without memory, the generated future continues moving after completion, producing a stop ratio above one. With memory-injection, post-completion motion drops to less than half of pre-completion motion. This indicates that episodic memory helps the video backbone represent not only what should happen next, but also when nothing further should happen. Please also refer to Figure \ref{fig:failure_case_} for failure modes without memory.

\paragraph{Dual injection ablation.}
\Cref{tab:libero} shows that memory is most effective when it shapes both the predictive representation and the action decoder. Pathway~1 alone reaches 28.5\% average success, indicating that history-aware backbone features already carry substantial task-progress information, but this information is not sufficient when the action decoder is not directly constrained by memory. Pathway~2 alone reaches 7.5\%, suggesting that exposing memory only at the action head cannot compensate for a history-agnostic backbone. In this case, the action decoder receives memory tokens while the world model still predicts futures from an incomplete episode state, creating a mismatch between the memory condition and visual features used for control. Our full model reaches 42.5\%, showing that memory must shape both the imagined future and downstream action distribution.

\section{Conclusion}
\label{sec:conclusion}
We presented \textbf{MemoryVAM}, an episodic memory module that is trained through the video generative objective and conditions both future prediction and action decoding via cross-attention. Across SVD UNet and Cosmos DiT backbones, the same memory mechanism improves long-horizon manipulation on LIBERO-Mem, raising average success from 5\% to 42.5\%, and transfers to real robots with 78.3\% success on counting tasks, 80.0\% on spatial recall, and 75.0\% on sequential tracking. More broadly, we reframe episodic memory as an integral component of the video world model, rather than an auxiliary policy module. By allowing episode history to shape both what the robot imagines and how the policy acts, our approach points toward memory systems whose supervision scales with video data and world-model training, rather than relying solely on action-labeled demonstrations.

\section{Limitations}
\label{sec:discussion}
\paragraph{Parallel vs.\ recurrent memory.}
The Recap Compressor trades direct access to retained history for memory and computation that grow with episode length. Very long deployments may therefore require truncation, subsampling, or an online recurrent state, which we leave as future work within the same dual-injection interface.

\paragraph{Scaling.}
Our method suggests a scaling opportunity because its memory module can be supervised by video prediction rather than only by demonstrations with action labels. Our experiments validate this mechanism on LIBERO-scale robot data and two video backbone families, but they do not establish how performance changes with substantially broader video pre-training or larger backbones. Studying this scaling behavior is therefore a promising direction for future work.

\clearpage
\bibliography{references}

@inproceedings{du2024vpp,
  title={Video Prediction Policy: A Generalist Robot Policy with Predictive Visual Representations},
  author={Hu, Yucheng and Guo, Yanjiang and Wang, Pengchao and Chen, Xiaoyu and Wang, Yen-Jen and Zhang, Jianke and Sreenath, Koushil and Lu, Chaochao and Chen, Jianyu},
  booktitle={Forty-second International Conference on Machine Learning}
}

@article{du2023unipi,
  title={Learning universal policies via text-guided video generation},
  author={Du, Yilun and Yang, Sherry and Dai, Bo and Dai, Hanjun and Nachum, Ofir and Tenenbaum, Josh and Schuurmans, Dale and Abbeel, Pieter},
  journal={Advances in neural information processing systems},
  volume={36},
  pages={9156--9172},
  year={2023}
}

@inproceedings{black2024susie,
  title={Zero-shot robotic manipulation with pre-trained image-editing diffusion models},
  author={Black, Kevin and Nakamoto, Mitsuhiko and Atreya, Pranav and Walke, Homer and Finn, Chelsea and Kumar, Aviral and Levine, Sergey},
  booktitle={International Conference on Learning Representations},
  volume={2024},
  pages={33431--33452},
  year={2024}
}

@inproceedings{wu2024gr1,
  title={Unleashing large-scale video generative pre-training for visual robot manipulation},
  author={Wu, Hongtao and Jing, Ya and Cheang, Chilam and Chen, Guangzeng and Xu, Jiafeng and Li, Xinghang and Liu, Minghuan and Li, Hang and Kong, Tao},
  booktitle={International Conference on Learning Representations},
  volume={2024},
  pages={10641--10662},
  year={2024}
}

@inproceedings{jaegle2021perceiver,
  title={Perceiver: General perception with iterative attention},
  author={Jaegle, Andrew and Gimeno, Felix and Brock, Andy and Vinyals, Oriol and Zisserman, Andrew and Carreira, Joao},
  booktitle={International conference on machine learning},
  pages={4651--4664},
  year={2021},
  organization={PMLR}
}

@article{ye2023ipadapter,
  title={Ip-adapter: Text compatible image prompt adapter for text-to-image diffusion models},
  author={Ye, Hu and Zhang, Jun and Liu, Sibo and Han, Xiao and Yang, Wei},
  journal={arXiv preprint arXiv:2308.06721},
  year={2023}
}

@inproceedings{he2024malmm,
  title={Ma-lmm: Memory-augmented large multimodal model for long-term video understanding},
  author={He, Bo and Li, Hengduo and Jang, Young Kyun and Jia, Menglin and Cao, Xuefei and Shah, Ashish and Shrivastava, Abhinav and Lim, Ser-Nam},
  booktitle={Proceedings of the IEEE/CVF conference on computer vision and pattern recognition},
  pages={13504--13514},
  year={2024}
}

@article{kerssies2026deltatok,
  title={A Frame is Worth One Token: Efficient Generative World Modeling with Delta Tokens},
  author={Kerssies, Tommie and Berton, Gabriele and He, Ju and Yu, Qihang and Ma, Wufei and de Geus, Daan and Dubbelman, Gijs and Chen, Liang-Chieh},
  journal={arXiv preprint arXiv:2604.04913},
  year={2026}
}

@article{liu2023libero,
  title={Libero: Benchmarking knowledge transfer for lifelong robot learning},
  author={Liu, Bo and Zhu, Yifeng and Gao, Chongkai and Feng, Yihao and Liu, Qiang and Zhu, Yuke and Stone, Peter},
  journal={Advances in Neural Information Processing Systems},
  volume={36},
  pages={44776--44791},
  year={2023}
}

@article{black2024pi0,
  title={$\pi_0$: A Vision-Language-Action Flow Model for General Robot Control},
  author={Black, Kevin and Brown, Noah and Driess, Danny and Esmail, Adnan and Equi, Michael and Finn, Chelsea and Fusai, Niccolo and Groom, Lachy and Hausman, Karol and Ichter, Brian and others},
  journal={arXiv preprint arXiv:2410.24164},
  year={2024}
}

@article{blattmann2023svd,
  title={Stable video diffusion: Scaling latent video diffusion models to large datasets},
  author={Blattmann, Andreas and Dockhorn, Tim and Kulal, Sumith and Mendelevitch, Daniel and Kilian, Maciej and Lorenz, Dominik and Levi, Yam and English, Zion and Voleti, Vikram and Letts, Adam and others},
  journal={arXiv preprint arXiv:2311.15127},
  year={2023}
}

@article{agarwal2025cosmos,
  title={Cosmos world foundation model platform for physical ai},
  author={Agarwal, Niket and Ali, Arslan and Bala, Maciej and Balaji, Yogesh and Barker, Erik and Cai, Tiffany and Chattopadhyay, Prithvijit and Chen, Yongxin and Cui, Yin and Ding, Yifan and others},
  journal={arXiv preprint arXiv:2501.03575},
  year={2025}
}

@article{kim2026cosmos,
  title={Cosmos policy: Fine-tuning video models for visuomotor control and planning},
  author={Kim, Moo Jin and Gao, Yihuai and Lin, Tsung-Yi and Lin, Yen-Chen and Ge, Yunhao and Lam, Grace and Liang, Percy and Song, Shuran and Liu, Ming-Yu and Finn, Chelsea and others},
  journal={arXiv preprint arXiv:2601.16163},
  year={2026}
}

@article{pai2025mimicvideo,
  title={mimic-video: Video-action models for generalizable robot control beyond vlas},
  author={Pai, Jonas and Achenbach, Liam and Montesinos, Victoriano and Forrai, Benedek and Mees, Oier and Nava, Elvis},
  journal={arXiv preprint arXiv:2512.15692},
  year={2025}
}

@inproceedings{bruce2024genie,
  title={Genie: Generative interactive environments},
  author={Bruce, Jake and Dennis, Michael D and Edwards, Ashley and Parker-Holder, Jack and Shi, Yuge and Hughes, Edward and Lai, Matthew and Mavalankar, Aditi and Steigerwald, Richie and Apps, Chris and others},
  booktitle={Forty-first International Conference on Machine Learning},
  year={2024}
}

@article{liao2025genieenvisioner,
  title={Genie envisioner: A unified world foundation platform for robotic manipulation},
  author={Liao, Yue and Zhou, Pengfei and Huang, Siyuan and Yang, Donglin and Chen, Shengcong and Jiang, Yuxin and Hu, Yue and Cai, Jingbin and Liu, Si and Luo, Jianlan and others},
  journal={arXiv preprint arXiv:2508.05635},
  year={2025}
}

@article{alayrac2022flamingo,
  title={Flamingo: a visual language model for few-shot learning},
  author={Alayrac, Jean-Baptiste and Donahue, Jeff and Luc, Pauline and Miech, Antoine and Barr, Iain and Hasson, Yana and Lenc, Karel and Mensch, Arthur and Millican, Katherine and Reynolds, Malcolm and others},
  journal={Advances in neural information processing systems},
  volume={35},
  pages={23716--23736},
  year={2022}
}

@article{jaegle2022perceiverio,
  title={Perceiver io: A general architecture for structured inputs \& outputs},
  author={Jaegle, Andrew and Borgeaud, Sebastian and Alayrac, Jean-Baptiste and Doersch, Carl and Ionescu, Catalin and Ding, David and Koppula, Skanda and Zoran, Daniel and Brock, Andrew and Shelhamer, Evan and others},
  journal={arXiv preprint arXiv:2107.14795},
  year={2021}
}

@inproceedings{li2023blip2,
  title={Blip-2: Bootstrapping language-image pre-training with frozen image encoders and large language models},
  author={Li, Junnan and Li, Dongxu and Savarese, Silvio and Hoi, Steven},
  booktitle={International conference on machine learning},
  pages={19730--19742},
  year={2023},
  organization={PMLR}
}

@article{chen2024diffusionforcing,
  title={Diffusion forcing: Next-token prediction meets full-sequence diffusion},
  author={Chen, Boyuan and Mart{\'\i} Mons{\'o}, Diego and Du, Yilun and Simchowitz, Max and Tedrake, Russ and Sitzmann, Vincent},
  journal={Advances in Neural Information Processing Systems},
  volume={37},
  pages={24081--24125},
  year={2024}
}

@article{huang2025selfforcing,
  title={Self forcing: Bridging the train-test gap in autoregressive video diffusion},
  author={Huang, Xun and Li, Zhengqi and He, Guande and Zhou, Mingyuan and Shechtman, Eli},
  journal={Advances in Neural Information Processing Systems},
  volume={38},
  pages={167283--167308},
  year={2026}
}

@article{robomme2026,
  title={Robomme: Benchmarking and understanding memory for robotic generalist policies},
  author={Dai, Yinpei and Fu, Hongze and Lee, Jayjun and Liu, Yuejiang and Zhang, Haoran and Yang, Jianing and Finn, Chelsea and Fazeli, Nima and Chai, Joyce},
  journal={arXiv preprint arXiv:2603.04639},
  year={2026}
}

@article{gao2026gatedmemorypolicy,
  title={Gated Memory Policy},
  author={Gao, Yihuai and Liu, Jinyun and Li, Shuang and Song, Shuran},
  journal={arXiv preprint arXiv:2604.18933},
  year={2026}
}

@article{memoryvla2025,
  title={Memoryvla: Perceptual-cognitive memory in vision-language-action models for robotic manipulation},
  author={Shi, Hao and Xie, Bin and Liu, Yingfei and Sun, Lin and Liu, Fengrong and Wang, Tiancai and Zhou, Erjin and Fan, Haoqiang and Zhang, Xiangyu and Huang, Gao},
  journal={arXiv preprint arXiv:2508.19236},
  year={2025}
}

@article{memer2026,
  title={Memer: Scaling up memory for robot control via experience retrieval},
  author={Sridhar, Ajay and Pan, Jennifer and Sharma, Satvik and Finn, Chelsea},
  journal={arXiv preprint arXiv:2510.20328},
  year={2025}
}

@inproceedings{slotssmv2,
  title={Rethinking progression of memory state in robotic manipulation: An object-centric perspective},
  author={Chung, Nhat and Hanyu, Taisei and Nguyen, Toan and Le, Huy and Bumgarner, Frederick and Nguyen, Duy Minh Ho and Vo, Khoa and Yamazaki, Kashu and Rainwater, Chase and Kieu, Tung and others},
  booktitle={Proceedings of the AAAI Conference on Artificial Intelligence},
  volume={40},
  number={5},
  pages={3407--3415},
  year={2026}
}

@article{torne2026mem,
  title={Mem: Multi-scale embodied memory for vision language action models},
  author={Torne, Marcel and Pertsch, Karl and Walke, Homer and Vedder, Kyle and Nair, Suraj and Ichter, Brian and Ren, Allen Z and Wang, Haohuan and Tang, Jiaming and Stachowicz, Kyle and others},
  journal={arXiv preprint arXiv:2603.03596},
  year={2026}
}

@article{hou2026world,
  title={World Model for Robot Learning: A Comprehensive Survey},
  author={Hou, Bohan and Li, Gen and Jia, Jindou and An, Tuo and Guo, Xinying and Leng, Sicong and Geng, Haoran and Ze, Yanjie and Harada, Tatsuya and Torr, Philip and others},
  journal={arXiv preprint arXiv:2605.00080},
  year={2026}
}

@inproceedings{radford2021learning,
  title={Learning transferable visual models from natural language supervision},
  author={Radford, Alec and Kim, Jong Wook and Hallacy, Chris and Ramesh, Aditya and Goh, Gabriel and Agarwal, Sandhini and Sastry, Girish and Askell, Amanda and Mishkin, Pamela and Clark, Jack and others},
  booktitle={International conference on machine learning},
  pages={8748--8763},
  year={2021},
  organization={PmLR}
}

\clearpage
\appendix
\begin{center}
    \LARGE \bf Supplementary Materials
\end{center}
\addtocontents{toc}{\protect\setcounter{tocdepth}{2}}
\tableofcontents
\section{Architecture Details}
\label{app:implementation}
\label{app:architecture_details}

\subsection{Recap Compressor and History Cache}
\label{app:recap_architecture}

The Recap Compressor implements the Perceiver-style resampler described in Sec.~\ref{sec:method:resampler}. Each cached CLIP visual embedding $\mathbf{x}_i \in \mathbb{R}^{d_{\text{clip}}}$ is projected to the memory dimension $d$ and augmented with temporal and visual-token positional encodings. The current implementation stores one global visual embedding per frame, so the visual-token position is constant while the temporal position indexes episode order. The $K$ learned queries are updated through $L$ cross-attention blocks with $N_{\text{head}}$ attention heads and feed-forward width multiplier $m_{\text{ffn}}$. This design gives the video backbone and action decoder a constant-size memory interface after the variable-length history has been compressed.

CLIP ViT-B/32 visual embeddings are precomputed for all episodes and stored in a cache. Each frame contributes one global visual embedding. Histories are padded or truncated to $T_{\max}$ embeddings during training and capped at $T_{\text{hist}}$ frames during evaluation, with a binary validity mask marking padded entries.

\subsection{Backbone Memory Interface}
\label{app:backbone_memory_interface}

For UNet video backbones, each text cross-attention layer receives a parallel memory branch with layer-specific key and value projections $\mathbf{W}_{\text{key},m}^{\text{mem}}$ and $\mathbf{W}_{\text{val},m}^{\text{mem}}$. The output is added to the original text-conditioned branch following decoupled cross-attention~\citep{ye2023ipadapter}. For DiT video backbones, memory tokens are appended to the conditioning matrix. In both cases, the Recap Compressor output $\mathbf{G}$ is unchanged, and only the injection interface depends on the backbone family.

\subsection{Action Decoder Memory Interface}
\label{app:action_memory_interface}

For token-input action decoders, Pathway~2 concatenates the memory tokens $\mathbf{G}$ with denoised video tokens $\mathbf{Z}_{\text{den},t}$. For cross-attention action transformers, projected memory tokens are appended to the visual conditioning sequence. These two instantiations implement the same conditional distribution in Sec.~\ref{sec:method:resampler}, where $\pi_\theta$ receives history either through $\mathbf{Z}_{\text{obs},t}$ or through an equivalent memory-augmented conditioning matrix.

\subsection{Cue Gate Architecture}
\label{app:progress_gate}

The Cue Gate is a lightweight classifier that fuses three information streams to estimate the episode-completion score $q_\omega(\mathbf{F}_{\text{vis},t}, \mathbf{G}, \ell) \in [0,1]$. Its dimensional constants are listed in \Cref{tab:architecture_parameters}.

\paragraph{Visual Stream.}
Backbone features $\mathbf{F}_{\text{vis},t}$ are pooled over spatial and temporal dimensions to produce $\bar{\mathbf{f}}_{\text{vis},t} \in \mathbb{R}^{d_{\text{vis}}}$. The visual branch projects this vector to the gate dimension,
\[
\mathbf{u}_{\text{vis}} =
\operatorname{Linear}_{d_{\text{vis}}\to d_{\text{gate}}}
\!\left(\operatorname{LayerNorm}(\bar{\mathbf{f}}_{\text{vis},t})\right).
\]

\paragraph{Memory Stream.}
Let $\mathbf{g}_k$ be the $k$th row of $\mathbf{G}$. The memory branch mean-pools the memory tokens and projects them to the same gate dimension,
\[
\bar{\mathbf{g}} = \frac{1}{K}\sum_{k=1}^{K}\mathbf{g}_k, \qquad
\mathbf{u}_{\text{mem}} =
\operatorname{Linear}_{d\to d_{\text{gate}}}
\!\left(\operatorname{LayerNorm}(\bar{\mathbf{g}})\right).
\]

\paragraph{Language Stream.}
The CLIP language embedding of the task instruction is denoted by $\mathbf{e}_{\text{lang}} \in \mathbb{R}^{d_{\text{lang}}}$. The language branch computes
\[
\mathbf{u}_{\text{lang}} =
\operatorname{Linear}_{d_{\text{lang}}\to d_{\text{gate}}}
\!\left(\operatorname{LayerNorm}(\mathbf{e}_{\text{lang}})\right).
\]

Let $\mathbf{u} = [\mathbf{u}_{\text{vis}};\,\mathbf{u}_{\text{mem}};\,\mathbf{u}_{\text{lang}}] \in \mathbb{R}^{3d_{\text{gate}}}$ denote the fused Cue Gate input. The completion score is
\begin{equation}
  q_\omega(\mathbf{F}_{\text{vis},t}, \mathbf{G}, \ell)
  = \sigma\!\left(
  \operatorname{Linear}_{d_{\text{gate}}\to 1}
  \!\left(
  \operatorname{GELU}\!\left(
  \operatorname{Linear}_{3d_{\text{gate}}\to d_{\text{gate}}}
  (\operatorname{LayerNorm}(\mathbf{u}))
  \right)\right)\right).
\end{equation}

\subsection{Reconstruction Decoder}
\label{app:reconstruction_decoder}

The reconstruction decoder uses $L_{\text{recon}}$ cross-attention layers followed by a feed-forward network to predict deltas between consecutive CLIP history embeddings. The decoder sends gradients directly into the Recap Compressor and is used only as an auxiliary training signal. The loss formulation is given in Appendix~\ref{app:recon}.

\section{Training Details}
\label{app:training_details}

\subsection{Stage 1 Video Model Training}
\label{app:stage1_training}

The memory-augmented video model is trained with
\[
\mathcal{L}
= \mathcal{L}_{\text{video}}
+ \lambda_{\text{recon}}\mathcal{L}_{\text{recon}}
+ \lambda_{\text{eos}}\mathcal{L}_{\text{eos}},
\]
where $\mathcal{L}_{\text{video}}$ is the denoising loss for SVD or the flow-matching loss for Cosmos, $\mathcal{L}_{\text{recon}}$ is the delta reconstruction loss in Appendix~\ref{app:recon}, and $\mathcal{L}_{\text{eos}}$ is the Cue Gate binary cross-entropy. Trainable components include the backbone attention layers, the Recap Compressor, memory injection projections, reconstruction decoder, and Cue Gate. The variational autoencoder (VAE) and CLIP text encoder remain frozen. To avoid early token collapse, we first train a memoryless video backbone and then resume from that checkpoint with memory modules added.

\subsection{Reconstruction Auxiliary Loss}
\label{app:recon}

Given memory tokens $\mathbf{G} \in \mathbb{R}^{K \times d}$ and the $M \leq T_{\max}$ valid CLIP history embeddings $\{\mathbf{x}_j\}_{j=0}^{M-1}$ for the current training sample, the reconstruction decoder predicts frame-to-frame feature deltas:
\begin{equation}
  \Delta\mathbf{x}_j = \mathbf{x}_j - \mathbf{x}_{j-1}, \qquad
  \widehat{\Delta\mathbf{x}}_j = f_{\text{recon}}(\mathbf{G},\; \mathbf{x}_{j-1}), \qquad
  \mathcal{L}_{\text{recon}} = \frac{1}{M-1} \sum_{j=1}^{M-1} \bigl\|\widehat{\Delta\mathbf{x}}_j - \Delta\mathbf{x}_j\bigr\|_2^2.
\end{equation}
The loss is evaluated only for samples with $M>1$ valid history embeddings.

Predicting deltas $\Delta\mathbf{x}_j$ rather than absolute features $\mathbf{x}_j$ forces the memory tokens to encode what changed between consecutive frames rather than memorizing static scene appearance. This design is inspired by DeltaTok~\citep{kerssies2026deltatok}, which shows that delta tokenization improves temporal coherence in video representations.

Without reconstruction, the Cue Gate produces noisy, low-confidence predictions that occasionally cross $q_\omega=0.5$ near the correct frame but cannot produce a sharp, high-confidence completion signal. As reported in \Cref{tab:eos_ablation}, overall timing error improves from 44.3 to 35.5 frames at $q_\omega \geq 0.5$. The gain is larger at $q_\omega \geq 0.9$ (62.0 $\to$ 28.8), confirming that the auxiliary loss improves calibration as well as timing accuracy.

\subsection{Cue Gate Supervision}
\label{app:cue_gate_training}

Training uses binary cross-entropy with $y_t=1$ for the final $w_{\text{eos}}$ frames of each episode and $y_t=0$ otherwise. Positive frames are upsampled by $m_{\text{eos}}$ within each batch to compensate for class imbalance. Gradients flow through the memory and language branches, while $\mathbf{F}_{\text{vis},t}$ is detached to prevent the Cue Gate loss from changing backbone representations.

Language conditioning is beneficial for the Cue Gate because the gate must evaluate which goal should be complete. In contrast, language-conditioned Recap queries hurt the task-agnostic history summary in the ablation in \Cref{sec:exp:ablation}.

\subsection{Stage 2 Action Policy Training}
\label{app:stage2_training}

The video backbone remains frozen for stability, but Pathway~1 remains active so the action policy trains on memory-conditioned backbone features. The Recap Compressor, memory injection projections, Cue Gate, Video Former, and GCDenoiser action head continue training with the action-prediction objective. For the VPP instantiation, the action head receives $K$ memory tokens together with $N_{\text{den}}$ denoised video tokens, giving $N_{\text{obs}}=K+N_{\text{den}}$ observation tokens.

\subsection{Inference}
\label{app:inference_details}

At test time, the current frame is encoded by CLIP and appended to the history cache. The Recap Compressor maps the retained history to $\mathbf{G}$, the video backbone predicts future frames and produces memory-conditioned features, and the action decoder predicts an action chunk from the resulting observation tokens. The Cue Gate score $q_\omega(\mathbf{F}_{\text{vis},t}, \mathbf{G}, \ell)$ terminates the rollout when it exceeds the threshold $\tau$ selected for the evaluation setting, or when the rollout reaches the task-specific maximum step count $t_{\max}$.

\section{Parameter Values}
\label{app:architecture_parameters}

\Cref{tab:architecture_parameters} lists the concrete values for the symbols used in Sec.~\ref{sec:method} and the appendix. Values marked as VPP-specific are used for the SVD UNet implementation; analogous DiT values are determined by the corresponding backbone hidden size. Case-specific choices such as the Cue Gate threshold and maximum rollout length are set by the corresponding evaluation protocol rather than treated as global constants.

\begin{table}[ht]
    \centering
    \small
    \setlength{\tabcolsep}{4pt}
    \caption{\textbf{Architecture, training, and evaluation parameter values.}}
    \label{tab:architecture_parameters}
    \resizebox{\linewidth}{!}{%
    \begin{tabular}{@{}llll@{}}
        \toprule
        Group & Symbol & Meaning & Value \\
        \midrule
        Video and policy & $r$ & Recent-frame observation context & 4 frames \\
         & $H_v$ & Video prediction chunk length & 16 frames \\
         & $H_a$ & Action chunk length & 8 actions \\
         & $R_{\text{img}}$ & Image resolution & 256 \\
        \addlinespace[2pt]\cmidrule(lr){2-4}\addlinespace[2pt]
        History cache & $d_{\text{clip}}$ & CLIP visual embedding dimension & 512 \\
         & $T_{\max}$ & Maximum training history embeddings & 400 \\
         & $T_{\text{hist}}$ & Maximum retained episode history & 600 frames \\
        \addlinespace[2pt]\cmidrule(lr){2-4}\addlinespace[2pt]
        Recap Compressor & $K$ & Memory tokens & 16 (VPP) / 32 (Cosmos) \\
         & $d$ & Memory hidden dimension & 384 / 1024 (Cosmos) \\
         & $L$ & Recap cross-attention layers & 4 \\
         & $N_{\text{head}}$ & Attention heads per Recap layer & 8 \\
         & $d_{\text{head}}$ & Attention head dimension & 64 \\
         & $m_{\text{ffn}}$ & Feed-forward width multiplier & 4 \\
         & $P_{\text{recap}}$ & Recap Compressor parameters & $\sim$4.7M / 32M (Cosmos) \\
        \addlinespace[2pt]\cmidrule(lr){2-4}\addlinespace[2pt]
        Reconstruction decoder & $L_{\text{recon}}$ & Reconstruction cross-attention layers & 1 \\
         & $P_{\text{recon}}$ & Reconstruction decoder parameters & $\sim$1.2M \\
        \addlinespace[2pt]\cmidrule(lr){2-4}\addlinespace[2pt]
        VPP action decoder & $N_{\text{den}}$ & Denoised video tokens & 224 \\
         & $N_{\text{obs}}$ & Action observation tokens & $K + N_{\text{den}}$ (240 for $K=16$) \\
        \addlinespace[2pt]\cmidrule(lr){2-4}\addlinespace[2pt]
        Cue Gate & $d_{\text{vis}}$ & Pooled backbone feature dimension (VPP) & 1280 \\
         & $d_{\text{lang}}$ & CLIP language embedding dimension & 1024 \\
         & $d_{\text{gate}}$ & Cue Gate branch dimension & 256 \\
         & $w_{\text{eos}}$ & Positive episode-boundary window & 10 frames \\
         & $m_{\text{eos}}$ & Positive-frame upsampling factor & 5 \\
        \addlinespace[2pt]\cmidrule(lr){2-4}\addlinespace[2pt]
        Training & $\lambda_{\text{recon}}$ & Reconstruction loss weight & 1.0 \\
         & $\lambda_{\text{eos}}$ & Cue Gate loss weight & 0.1 \\
         & $\eta_{\text{video}}$ & Stage~1 base learning rate & $5 \times 10^{-6}$ \\
         & $\eta_{\text{mem}}$ & Stage~1 memory learning rate & $5 \times 10^{-5}$ \\
         & $m_{\text{mem}}$ & Memory learning-rate multiplier & $10\times$ \\
         & $B_1$ & Stage~1 effective batch size & 32 \\
         & $S_{\text{base}}$ & Memoryless warm-start checkpoint & $270{,}000$ steps \\
         & $S_1$ & Total Stage~1 training & $500{,}000$ steps \\
         & $\sigma_{\text{clip}}$ & CLIP embedding noise std. & 0.1 \\
         & $\eta_{\text{act}}$ & Stage~2 learning rate & $10^{-4}$ \\
         & $B_2$ & Stage~2 batch size & 16 \\
         & $N_{\text{DDIM}}$ & DDIM denoising steps for action decoding & 10 \\
        \addlinespace[2pt]\cmidrule(lr){2-4}\addlinespace[2pt]
        Stop ratio & $s_{\text{stop}}$ & Grayscale evaluation resolution & 64 \\
         & $w_{\text{pre}}$ & Pre-completion motion window & 48 frames \\
         & $w_{\text{post}}$ & Post-completion motion window & 72 frames \\
         & $\epsilon$ & Stop-ratio numerical stabilizer & $10^{-8}$ \\
        \bottomrule
    \end{tabular}
    }
\end{table}

\section{Evaluation Definitions}
\label{app:evaluation_definitions}

\subsection{Stop Ratio}
\label{app:stop}

The \emph{stop ratio} measures whether a generated video physically halts after the task is completed. We compute it as the ratio of post-completion to pre-completion motion energy using the constants in \Cref{tab:architecture_parameters}.

Each generated frame is converted to grayscale at $s_{\text{stop}} \times s_{\text{stop}}$ resolution and denoted by $\mathbf{Y}_t$. Per-frame motion is
\[
\delta_t = \operatorname{mean}\!\left(|\mathbf{Y}_{t+1} - \mathbf{Y}_t|\right).
\]
Let $t^*$ denote the ground-truth completion frame. We define
\[
E_{\text{pre}} = \frac{1}{w_{\text{pre}}}\sum_{u=t^*-w_{\text{pre}}}^{t^*-1} \delta_u, \qquad
E_{\text{post}} = \frac{1}{w_{\text{post}}}\sum_{u=t^*}^{t^*+w_{\text{post}}-1} \delta_u, \qquad
\text{stop ratio} = \frac{E_{\text{post}}}{E_{\text{pre}} + \epsilon}.
\]
A stop ratio below 1 indicates that motion decreases after completion, while a ratio above 1 indicates continued spurious motion. \Cref{tab:video_stop} reports the corresponding termination results.

\begin{table}[t]
\centering
\makebox[\linewidth][c]{%
\begin{subtable}[t]{0.508\linewidth}
    \centering
    \caption{}
    \label{tab:vidpred}
    \scriptsize
    \setlength{\tabcolsep}{1.2pt}
    \renewcommand{\arraystretch}{0.96}
    \begin{tabular*}{\linewidth}{@{\extracolsep{\fill}}lcccccc@{}}
        \toprule
        & \multicolumn{3}{c}{Single $H_v$} & \multicolumn{3}{c}{Iter. FVD$\downarrow$} \\
        \cmidrule(lr){2-4}\cmidrule(lr){5-7}
        Method & SSIM$\uparrow$ & LPIPS$\downarrow$ & FVD$\downarrow$ & $2H_v$ & $4H_v$ & $8H_v$ \\
        \midrule
        VPP & 0.7711 & 0.1673 & 43.36 & 39.40 & 33.10 & 37.81 \\
        \textbf{Ours (UNet)} & \textbf{0.8094} & \textbf{0.1145} & \textbf{18.32} & \textbf{17.01} & \textbf{17.11} & \textbf{19.84} \\
        \midrule
        Cosmos-RF & 0.8744 & 0.0897 & 22.54 & \textbf{23.78} & \textbf{21.23} & 26.11 \\
        \textbf{Ours (DiT)} & \textbf{0.8752} & \textbf{0.0781} & \textbf{18.65} & 25.37 & 22.06 & \textbf{16.24} \\
        \bottomrule
    \end{tabular*}
\end{subtable}\hspace{0.22em}%
\begin{subtable}[t]{0.315\linewidth}
    \centering
    \caption{}
    \label{tab:eos_ablation}
    \scriptsize
    \setlength{\tabcolsep}{1.5pt}
    \renewcommand{\arraystretch}{0.96}
    \begin{tabular*}{\linewidth}{@{\extracolsep{\fill}}llcccc@{}}
        \toprule
        \multicolumn{6}{c}{$|\Delta|$ (frames$\downarrow$)} \\
        \cmidrule(lr){1-6}
        Var. & Thr. & 1$\times$ & 3$\times$ & 5$\times$ & All \\
        \midrule
        w/o & .5 & \textbf{29.5} & 24.5 & 79.1 & 44.3 \\
        \textbf{w/} & .5 & 37.2 & \textbf{14.8} & \textbf{54.6} & \textbf{35.5} \\
        \midrule
        w/o & .9 & 72.3 & 33.0 & 80.7 & 62.0 \\
        \textbf{w/} & .9 & \textbf{21.0} & \textbf{14.6} & \textbf{50.8} & \textbf{28.8} \\
        \bottomrule
    \end{tabular*}
\end{subtable}\hspace{0.22em}%
\begin{subtable}[t]{0.165\linewidth}
    \centering
    \caption{}
    \label{tab:video_stop}
    \scriptsize
    \setlength{\tabcolsep}{1.4pt}
    \renewcommand{\arraystretch}{0.96}
    \begin{tabular*}{\linewidth}{@{\extracolsep{\fill}}lcc@{}}
        \toprule
        \multicolumn{3}{c}{Ratio$\downarrow$} \\
        \cmidrule(lr){1-3}
        Var. & Ratio & Energy \\
        \midrule
        No mem & 1.11 & 4.10 \\
        \midrule
        \multicolumn{3}{c}{\phantom{0}} \\
        w/o & 0.53 & 1.67 \\
        \textbf{w/} & \textbf{0.46} & \textbf{1.00} \\
        \bottomrule
    \end{tabular*}
\end{subtable}
}
\vspace{0.1em}
\caption{\textbf{Generation and Completion Quality} on LIBERO-Mem~\citep{slotssmv2}. (\subref{tab:vidpred})~video prediction fidelity, (\subref{tab:eos_ablation})~completion-timing error in frames, and (\subref{tab:video_stop})~post-completion motion. w/ and w/o denote with and without the reconstruction auxiliary loss.}
\label{tab:gen_quality}
\end{table}

\section{Memory Analysis Details}
\label{app:analysis}

This section provides diagnostic analysis of how the memory tokens encode task progress.

\paragraph{Cue Gate Probability over Time.}
We analyze the Cue Gate output $q_\omega$ over the course of a $3\times$ repetition task.
The curve remains near zero during inter-repetition transitions and crosses the high-confidence threshold $q_\omega>0.9$ only when the final repetition is completed.
During intermediate repetitions, the gate produces small transient bumps in the range $0.1$--$0.3$ that do not trigger termination, demonstrating that the model counts task progress rather than responding to any single placement event.

\paragraph{Decision-Critical Frame Comparison.}
The third placement in a $3\times$ task and the third placement in a $5\times$ task look nearly identical in raw visual features, with cosine similarity between CLIP frame embeddings exceeding 0.94.
Yet the Cue Gate assigns $q_\omega > 0.9$ in the $3\times$ case and $q_\omega < 0.2$ in the $5\times$ case.
This confirms that the gate's decision depends on the full history encoded in the memory tokens, including the count of prior repetitions, rather than on the current visual observation alone.

\paragraph{Shuffled Memory Control.}
To verify that temporal order matters, we shuffle the CLIP history sequence before passing it through the Recap Compressor at inference time.
Shuffling destroys the chronological signal while preserving the same set of visual features.
Timing accuracy degrades substantially, confirming that the Recap Compressor exploits temporal ordering to encode task progress and does not rely on bag-of-frames statistics or other spurious cues.

\section{Real-Robot Experimental Setup}
\label{app:real_robot_setup}

\begin{figure}[t]
    \centering
    \includegraphics[width=0.95\linewidth]{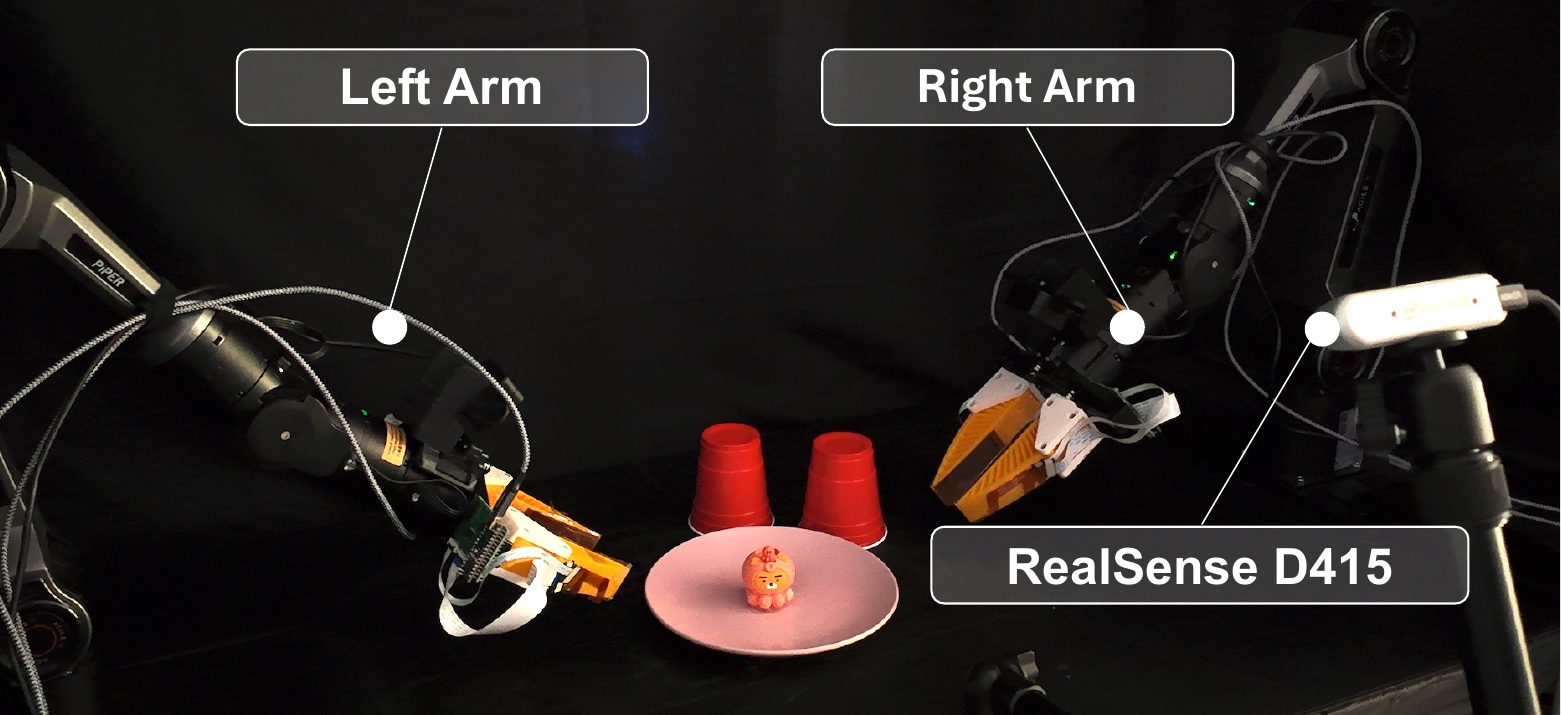}
    \caption{\textbf{Real-Robot Setup for MemoryVAM Evaluation.} The setup includes PiPER tabletop manipulation arms, the object set used in the real-robot memory tasks, and a fixed Intel RealSense D415 camera for capturing visual observations.}
    \label{fig:robot_setup}
\end{figure}

\Cref{fig:robot_setup} shows the physical setup used for the real-robot evaluations in Sec.~\ref{sec:exp:real}. The hardware uses PiPER tabletop manipulation arms in task-dependent configurations, with single-arm control for the single-arm evaluations and a bimanual setup for the bimanual evaluation. A fixed Intel RealSense D415 camera is mounted above the workspace to capture visual observations for policy execution.

\begin{figure}[t]
    \centering
    \includegraphics[width=0.95\linewidth]{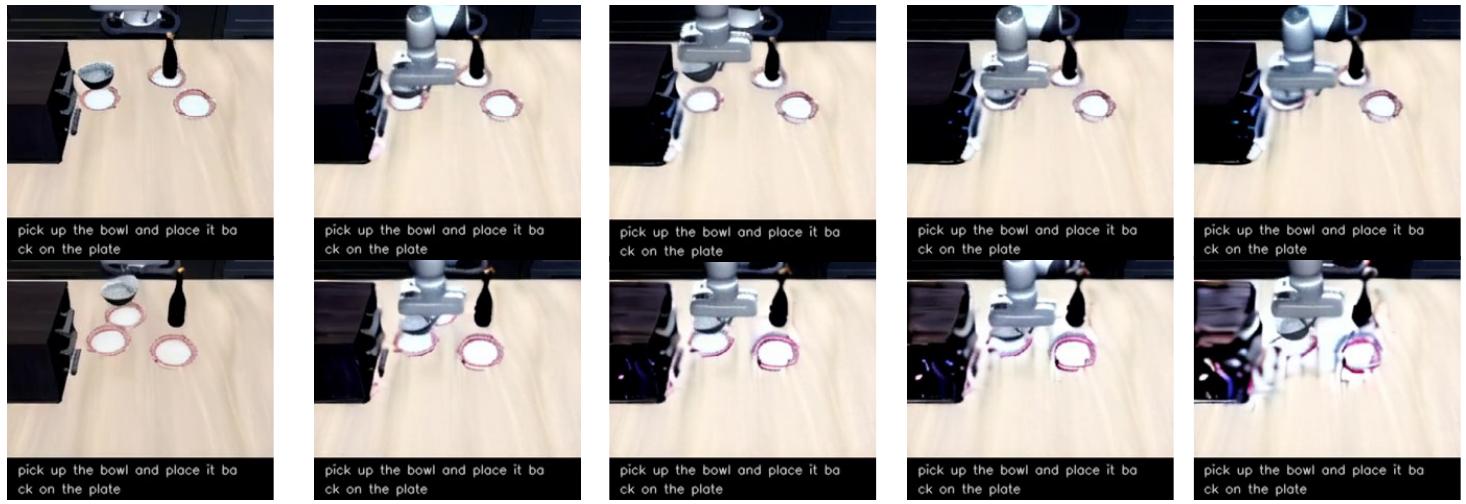}
    \caption{\textbf{Comparison of Video Generation Quality}:
    Video generation results w/ and w/o memory. \emph{Top}: Video generation with memory can accurately terminate upon task completion while maintaining stable long-horizon visual consistency.
\emph{Bottom}: Video generation without memory exhibits significant temporal fluctuations, progressive degradation over time, and fails to reliably determine when the task has ended.}
    \label{fig:failure_case_}
\end{figure}

\section{Video Generation Quality Results}
\Cref{fig:failure_case_} compares video generation results from models w/ and w/o memory.

\end{document}